\documentclass[runningheads]{llncs}

\usepackage[utf8]{inputenc}
\usepackage[english]{babel}
\usepackage{xspace}
\usepackage{graphicx,float}
\usepackage[font=small,labelfont=bf]{caption}
\usepackage[labelformat=simple]{subcaption}
\usepackage[title]{appendix}
\usepackage{lineno}
\usepackage{csquotes}
\usepackage[backend=biber,style=numeric,sorting=none]{biblatex}
\usepackage{amsmath,amsbsy,bm,amssymb,mathtools,dsfont}
\usepackage{etoolbox}
\usepackage{tikz,pgfplots}
\usepackage{enumitem}
\usepackage{tabu,makecell,diagbox,booktabs}
\usepackage[group-minimum-digits=3,separate-uncertainty]{siunitx}
\usepackage[hidelinks]{hyperref}
\usepackage{orcidlink}
\usepackage{cleveref}
\usepackage[top=3cm,bottom=3cm,left=3cm,right=3cm]{geometry}

\NewBibliographyString{toappear}
\DefineBibliographyStrings{english}{%
	toappear = {submitted to},
}
\renewbibmacro*{in:}{%
	\iffieldundef{pubstate}
	{\bibstring{in}}
	{\printfield{pubstate}%
		\setunit{\addspace}%
		\clearfield{pubstate}}%
	\printtext{%
		\intitlepunct}%
}

\definecolor{cupcol}{rgb}{.0,.0,.0}
\definecolor{refcol}{rgb}{.0,.0,.0}
\definecolor{inpcol}{rgb}{.8,.0,.0}
\definecolor{outcol}{rgb}{.0,.0,.8}
\definecolor{prucol}{rgb}{.0,.8,.0}

\crefname{figure}{Fig.}{Figs.}
\Crefname{figure}{Fig.}{Figs.}
\crefname{table}{Table}{Tables}
\Crefname{table}{Table}{Tables}
\crefname{equation}{Eq.}{Eqs.}
\Crefname{equation}{Eq.}{Eqs.}

\pgfplotsset{compat=newest}

\newcommand{\ie}{i.\,e.\xspace}
\newcommand{\eg}{e.\,g.\xspace}
\newcommand{\cf}{cf.\xspace}

\newcommand{\quoting}[1]{``#1''}
\newcommand{\cnum}[1]{\color{green!35!black}\textbf{\num{#1}}}
\newcommand{\rnum}[1]{\color{red!35!black}\textbf{\num{#1}}}

\newcommand{\withsep}{\hspace{.85cm}\text{with}\hspace{.85cm}}

\newcommand{\inlinecolorbar}[3]{%
\tikz[overlay,yshift=#3]{\begin{axis}[hide axis,scale only axis,height=0pt,width=0pt,colormap=#1,colormap access=#2,colorbar horizontal,colorbar style={height=.25cm,width=1cm,xtick=\empty}]\end{axis}}%
\hspace{1.1cm}}

\newcommand{\rhotcolorbar}{\inlinecolorbar{{reverse hot}{indices of colormap={0,...,\pgfplotscolormaplastindexof{hot} of hot}}}{map}{3.9ex}}

\newcommand{\cupnet}{\texttt{CupNet}\xspace}
\newcommand{\regnet}{\texttt{RefNet}\xspace}

\begin{document}

\title{CupNet -- Pruning a network for geometric data} 

\author{Raoul Heese\inst{1,2,}\thanks{\email{raoul.heese@itwm.fraunhofer.de}}\orcidlink{0000-0001-7479-3339} \and%
	Lukas Morand\inst{3}\orcidlink{0000-0002-8566-7642} \and%
	Dirk Helm\inst{3}\orcidlink{0000-0002-5398-5443} \and%
	Michael Bortz\inst{1,2}\orcidlink{0000-0001-8169-2907}}
\authorrunning{Heese, Morand, Helm, Bortz}
\institute{%
	Fraunhofer Center for Machine Learning \and%
	Fraunhofer-Institut f{\"u}r Techno- und Wirtschaftsmathematik ITWM, Fraunhofer-Platz 1, 67663 Kaiserslautern, Germany \and%
	Fraunhofer-Institut f{\"u}r Werkstoffmechanik IWM, W{\"o}hlerstr. 11, 79108 Freiburg, Germany%
	}

\maketitle

\begin{abstract}
	Using data from a simulated cup drawing process, we demonstrate how the inherent geometrical structure of cup meshes can be used to effectively prune an artificial neural network in a straightforward way.
\end{abstract}

\section{Introduction}
The optimization of production processes can benefit from machine learning methods that incorporate domain knowledge and data from numerical simulations \cite{rueden2020}. Typically, such methods aim to model relations between process parameters and the resulting product. In this manuscript, we consider an example from the field of deep drawing, a sheet metal forming process in which a sheet metal blank is drawn into a forming die by mechanical action.\par
Specifically, we study the prediction of product geometries in a cup drawing process based on data from finite element simulations \cite{morand2020}. For each simulation, we choose randomized process and material parameters $\mathbf{p} \in \mathbb{R}^{k}$ with $k \equiv \num{9}$ and observe the resulting geometry as a set of $m \equiv \num{1979}$ mesh coordinates $\mathbf{x} \in \mathbb{R}^d$ with $d \equiv 3 m = \num{5937}$. Thus, the machine learning task is to predict
\begin{align} \label{eqn:model}
	\mathbf{\hat{x}}(\mathbf{p}) : \mathbb{R}^{k} \longmapsto \mathbb{R}^d
\end{align}
based on the generated data. Such a predictive regression model can be considered as a short-cut for the actual simulation. In contrast to the simulation, it is faster and always numerically stable and therefore particularly suitable to solve optimization problems. On the other hand, the model predictions are less accurate than the simulation results, which corresponds to a trade-off between calculation speed and outcome precision.\par
\begin{figure}[tbh]
	\begin{center}
		\begin{subfigure}[t]{.28\linewidth}
			\centering\includegraphics[scale=1]{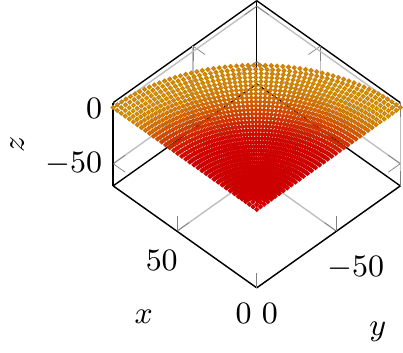}
			\caption{Before deformation.} \label{fig:cup:before}
		\end{subfigure}\hfill%
		\begin{subfigure}[t]{.23\linewidth}
			\centering\includegraphics[scale=1]{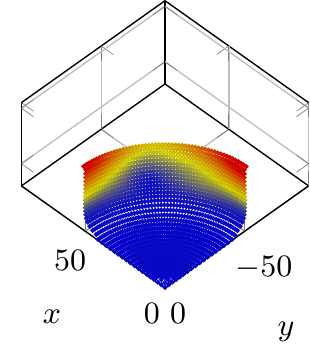}
			\caption{Good cup.} \label{fig:cup:ok}
		\end{subfigure}\hfill%
		\begin{subfigure}[t]{.23\linewidth}
			\centering\includegraphics[scale=1]{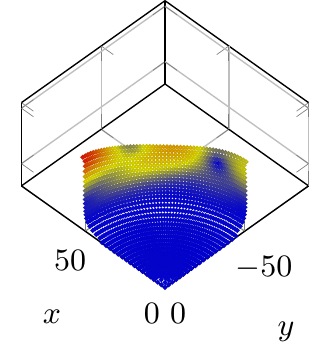}
			\caption{Defect cup.} \label{fig:cup:damaged}
		\end{subfigure}\hfill%
		\begin{subfigure}[t]{.23\linewidth}
			\centering\includegraphics[scale=1]{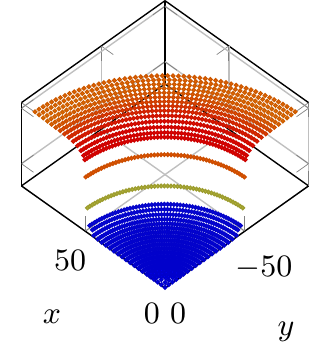}
			\caption{Cracked cup.} \label{fig:cup:broken}
		\end{subfigure}	
	\end{center}
	\caption{Typical cup geometries, each consisting of $m$ points. For reasons of symmetry, it is sufficient to simulate the deformation of a quarter cup segment instead of the full cup. The colors indicate the distance of each point to the reference mesh, consisting of the average coordinates of all good cups. We use a different color scale for each subfigure: 0 \protect\rhotcolorbar \subref{fig:cup:before} 70, \subref{fig:cup:ok} 2, \subref{fig:cup:damaged} 6, \subref{fig:cup:broken} 49.} \label{fig:cup}
\end{figure}

The choice of parameters affects the resulting cup quality in the sense that we can infer good, defect and cracked cups (indicated by strong deformations) from the mesh geometries. In total, we ran \num{10000} simulations, of which two failed (for numerical reasons). Of the remaining \num{9998} parameter combinations, \num{3991} lead to good cups, \num{5075} lead to defect cups and \num{932} cause cracked cups, \cf \cref{fig:cup}.

\section{Method} \label{sec:method}
We propose two artificial neural networks to model \cref{eqn:model}. Our first network architecture, which we call \cupnet, particularly takes the geometrical structure of the data into account to effectively prune the network weights. Pruning is a technique that helps in the development of smaller and more efficient networks, see, \eg, Refs.~\cite{liu2018,blalock2020} and references therein. That means, instead of changing the loss function as in \eg, Ref.~\cite{heese2019}, we use expert knowledge to change the network architecture itself.\par
The proposed network consists of an input layer of size $k$ (\ie, it contains $k$ units), which is fully connected to an initial layer of size $d$, which we call \emph{frame}. We split the frame layer into three evenly sized segments (\ie, one for each dimension denoted by $x$, $y$, and $z$, \cf \cref{fig:cup}), which are each connected to the following layers in a special way. Specifically, we chose the forward pass
\begin{align} \label{eqn:forward}
\mathbf{o} \equiv A( [\mathbf{C}(\alpha) \odot \mathbf{W}]^\intercal \mathbf{i} + \mathbf{b}).
\end{align}
Here $\mathbf{i} \in \mathbb{R}^{m \times 1}$ and $\mathbf{o} \in \mathbb{R}^{m \times 1}$ represent the layer inputs and outputs, whereas $\mathbf{W} \in \mathbb{R}^{m \times m}$ and $\mathbf{b} \in \mathbb{R}^{m \times 1}$ stand for the (trainable) layer weights and biases, and $A(\cdot)$ represents the activation function. The symbol $\odot$ denotes the Hadamard product (element-wise multiplication). Moreover, we have introduced the symmetric pruning matrix $\mathbf{C}(\alpha) \in \{0,1\}^{m \times m}$ with elements
\begin{align} \label{eqn:C}
\mathbf{C}_{ij}(\alpha) \equiv \begin{cases} 0 & \text{if}\,\, \mathbf{D}_{ij} > \alpha \\ 1 & \text{if}\,\, \mathbf{D}_{ij} \leq \alpha \end{cases} \withsep \mathbf{D}_{ij} \equiv \vert\vert \mathbf{x'_i} - \mathbf{x'_j} \vert\vert_2 = \operatorname{const.}
\end{align}
for $i,j = 1,\dots,m$. It is based on the symmetric distance matrix $\mathbf{D} \in \mathbb{R}_{\geq 0}^{m \times m}$, which contains the euclidean distances $\vert\vert \cdot \vert\vert_2$ between different mesh points $\mathbf{x'}$ of the undeformed geometry, \cref{fig:cup:before}. Thus, the pruning matrix removes the influence of all weights for which the corresponding mesh points of the undeformed geometry have a distance beyond the user-defined pruning threshold $\alpha \geq 0$.\par
\begin{figure}[tbh]
	\begin{center}
		\begin{subfigure}[t]{.45\linewidth}
			\includegraphics[scale=1]{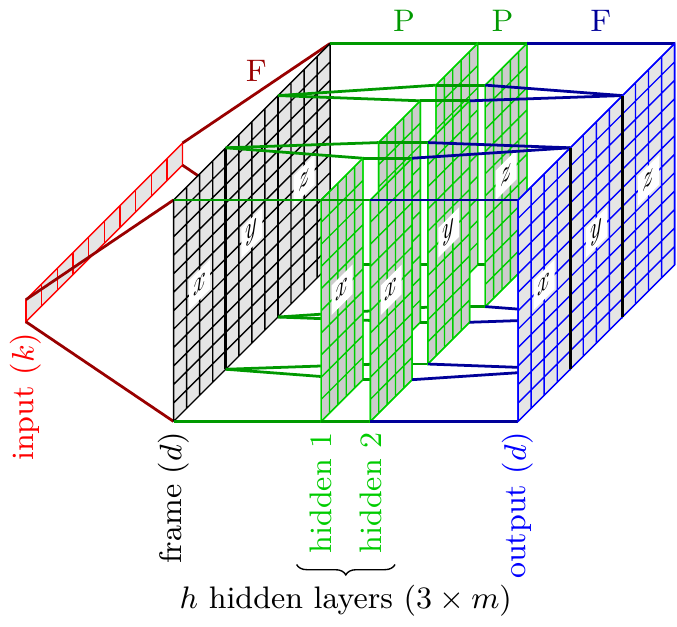}
			\caption{Our proposed \cupnet with $n_{\mathrm{cup}}(h,\alpha)$ trainable parameters, \cref{eqn:ncup}, and dimensions $x$, $y$, and $z$.} \label{fig:net:cup}
		\end{subfigure}
		\hfill%
		\begin{subfigure}[t]{.45\linewidth}
			\includegraphics[scale=1]{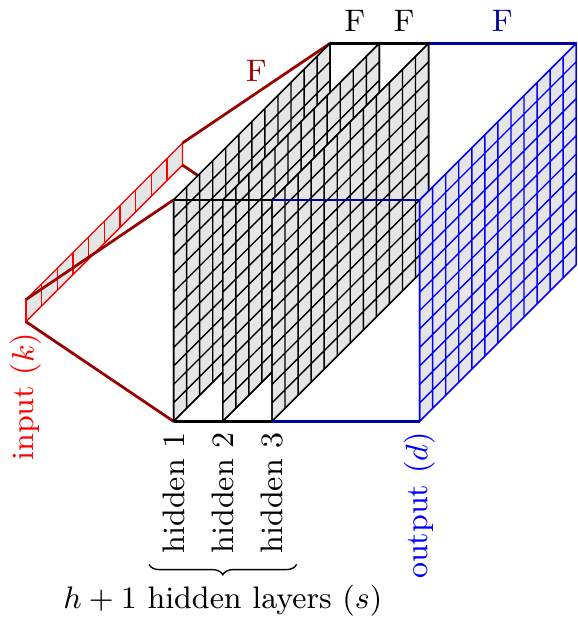}
			\caption{The reference network \regnet with $n_{\mathrm{ref}}(h,s)$ trainable parameters, \cref{eqn:nreg}.}\label{fig:net:reg}
		\end{subfigure}	
	\end{center}
	\caption{Network architectures used to model \cref{eqn:model}. The $k$-dimensional simulation parameters $\color{inpcol}\mathbf{p}$ constitute the input, whereas the output corresponds to the $d$-dimensional mesh ${\color{outcol}\mathbf{\hat{x}}}({\color{inpcol}\mathbf{p}})$. Both architectures consist of a sequence of layers (number of units in brackets), which are fully connected (F) or partially connected, \ie, pruned ({\color{prucol}P}). We apply a dropout after every inner layer for regularization.} \label{fig:net}
\end{figure}

This special layer configuration is repeated $h$ times and concludes with a fully-connected last layer merging the three previously splitted segments into the output layer. Summarized, we use the inherent geometrical structure of the data to prune a fully connected network in such a way that correlations between spatially related mesh points are preserved. The complete architecture is sketched in \cref{fig:net:cup}.\par

As a reference we also use a second architecture, which we call \regnet. It has a similar structure and complexity as the \cupnet, but does not take advantage of the geometrical structure of the data. Specifically, it consists of an input layer of size $k$, $h+1$ hidden layers of size $s$ and an output layer of size $m$, all of which are fully connected. In order to obtain a comparable complexity of the two architectures we choose $s$ in such a way that the number of trainable parameters is as close as possible. For that, we first determine the number of trainable parameters for the \cupnet architecture
\begin{align} \label{eqn:ncup}
n_{\mathrm{cup}}(h,\alpha) = (k \cdot d + d) +  3 h (c(\alpha) + m) + 3 (m^2 + m) \withsep c(\alpha) \equiv \sum_{i,j=1}^m \mathbf{C}_{ij}(\alpha).
\end{align}
On the other hand, the number of trainable parameters for the \regnet architecture reads
\begin{align} \label{eqn:nreg}
n_{\mathrm{ref}}(h,s) = (k \cdot s + s) +  h (s^2 + s) + (s \cdot d + d)
\end{align}
so that the condition $n_{\mathrm{ref}}(h,s) \overset{!}{=} n_{\mathrm{cup}}(h,\alpha)$ leads to $s(h,\alpha) = (-u+\sqrt{u^2 - 4 h (d-n_{\mathrm{cup}}(h,\alpha))}) / (2 h)$, where $u \equiv k+h+d+1$. We set the layer size $s$ to the ceiling of this value for a given depth $h$ and a given pruning threshold $\alpha$. The complete \regnet architecture is sketched in \cref{fig:net:reg}.\par
For both networks the activation function is chosen to be a ReLU for all but the output layer for which we use a linear activation. As a regularizer we apply a dropout after every inner layer, where we randomly set $\num{20}\%$ of the units to zero.

\section{Experiments}
We test the performance of both network architectures, which we have implemented using Ref.~\cite{tensorflow}. For this purpose we split the standardized data into a training set with \num{9000} elements and a test set with \num{998} elements stratified according to the three cup classes of good cups, defect cups, and cracked cups. We repeat this approach for \num{10} benchmark runs with different data splittings and different random seeds for the network initialization. For the training we use an Adam optimizer and a mean squared error for the loss. As a result, we obtain the $R^2$ scores shown in \cref{tab:results}.\par

\begin{table}[t]
	\centering
	\caption{$R^2$ score means and standard deviations over all 10 benchmark runs for different values of the network depth $h$ and the pruning threshold $\alpha$. The best mean results are highlighted in colored bold.}\label{tab:results}
	\resizebox{\linewidth}{!}{%
	\begin{tabular}{l|c|c|c|c|c|c}\hline
		%
		\multicolumn{1}{l|}{\textbf{network}} & $\alpha=\num{1.0}$       			 & $\alpha=\num{2.5}$           		 & $\alpha=\num{5.0}$            		 &%
		$\alpha=\num{10.0}$                   & $\alpha=\num{25.0}$                   & $\alpha=\num{50.0}$                   \\ \hline \hline
		{\color{cupcol}\cupnet $(h=\num{1})$} & {\color{cupcol}\cnum{0.822 +- 0.042}} & {\color{cupcol}\cnum{0.823 +- 0.052}} & {\color{cupcol}\cnum{0.829 +- 0.040}} & {\color{cupcol}\cnum{0.832 +- 0.084}} & {\color{cupcol}\cnum{0.839 +- 0.044}} & {\color{cupcol}\cnum{0.819 +- 0.072}} \\ \hline       
		{\color{refcol}\regnet $(h=\num{1})$} & {\color{refcol}\num {0.818 +- 0.047}} & {\color{refcol}\num {0.818 +- 0.066}} & {\color{refcol}\num {0.815 +- 0.066}} & {\color{refcol}\num {0.817 +- 0.045}} & {\color{refcol}\num {0.818 +- 0.056}} & {\color{refcol}\num {0.817 +- 0.053}} \\ \hline\hline
		{\color{cupcol}\cupnet $(h=\num{2})$} & {\color{cupcol}\cnum{0.821 +- 0.038}} & {\color{cupcol}\cnum{0.850 +- 0.039}} & {\color{cupcol}\cnum{0.862 +- 0.046}} & {\color{cupcol}\cnum{0.851 +- 0.057}} & {\color{cupcol}\cnum{0.847 +- 0.047}} & {\color{cupcol}\cnum{0.831 +- 0.045}} \\ \hline       
		{\color{refcol}\regnet $(h=\num{2})$} & {\color{refcol}\num {0.804 +- 0.050}} & {\color{refcol}\num {0.813 +- 0.061}} & {\color{refcol}\num {0.809 +- 0.048}} & {\color{refcol}\num {0.821 +- 0.036}} & {\color{refcol}\num {0.803 +- 0.037}} & {\color{refcol}\num {0.800 +- 0.070}} \\ \hline\hline
		{\color{cupcol}\cupnet $(h=\num{3})$} & {\color{cupcol}\cnum{0.830 +- 0.039}} & {\color{cupcol}\cnum{0.849 +- 0.039}} & {\color{cupcol}\cnum{0.850 +- 0.053}} & {\color{cupcol}\cnum{0.861 +- 0.056}} & {\color{cupcol}\cnum{0.839 +- 0.053}} & {\color{cupcol}\cnum{0.829 +- 0.043}} \\ \hline       
		{\color{refcol}\regnet $(h=\num{3})$} & {\color{refcol}\num {0.811 +- 0.043}} & {\color{refcol}\num {0.789 +- 0.085}} & {\color{refcol}\num {0.826 +- 0.042}} & {\color{refcol}\num {0.802 +- 0.064}} & {\color{refcol}\num {0.810 +- 0.045}} & {\color{refcol}\num {0.827 +- 0.029}} \\ \hline\hline
		{\color{cupcol}\cupnet $(h=\num{4})$} & {\color{cupcol}\cnum{0.830 +- 0.040}} & {\color{cupcol}\cnum{0.838 +- 0.050}} & {\color{cupcol}\cnum{0.845 +- 0.064}} & {\color{cupcol}\cnum{0.849 +- 0.035}} & {\color{cupcol}\cnum{0.817 +- 0.114}} & {\color{cupcol}\cnum{0.806 +- 0.034}} \\ \hline       
		{\color{refcol}\regnet $(h=\num{4})$} & {\color{refcol}\num {0.797 +- 0.055}} & {\color{refcol}\num {0.792 +- 0.044}} & {\color{refcol}\num {0.800 +- 0.052}} & {\color{refcol}\num {0.808 +- 0.041}} & {\color{refcol}\num {0.779 +- 0.051}} & {\color{refcol}\num {0.768 +- 0.068}} \\ \hline\hline
		{\color{cupcol}\cupnet $(h=\num{5})$} & {\color{cupcol}\num {0.478 +- 0.292}} & {\color{cupcol}\cnum{0.835 +- 0.064}} & {\color{cupcol}\cnum{0.838 +- 0.048}} & {\color{cupcol}\cnum{0.839 +- 0.037}} & {\color{cupcol}\cnum{0.832 +- 0.036}} & {\color{cupcol}\cnum{0.805 +- 0.033}} \\ \hline       
		{\color{refcol}\regnet $(h=\num{5})$} & {\color{refcol}\rnum{0.762 +- 0.060}} & {\color{refcol}\num {0.774 +- 0.041}} & {\color{refcol}\num {0.789 +- 0.038}} & {\color{refcol}\num {0.767 +- 0.050}} & {\color{refcol}\num {0.778 +- 0.042}} & {\color{refcol}\num {0.691 +- 0.126}} \\ \hline\hline
		{\color{cupcol}\cupnet $(h=\num{6})$} & {\color{cupcol}\num {0.052 +- 0.108}} & {\color{cupcol}\cnum{0.832 +- 0.042}} & {\color{cupcol}\cnum{0.840 +- 0.046}} & {\color{cupcol}\cnum{0.832 +- 0.103}} & {\color{cupcol}\cnum{0.832 +- 0.045}} & {\color{cupcol}\cnum{0.787 +- 0.080}} \\ \hline       
		{\color{refcol}\regnet $(h=\num{6})$} & {\color{refcol}\rnum{0.780 +- 0.041}} & {\color{refcol}\num {0.785 +- 0.039}} & {\color{refcol}\num {0.769 +- 0.049}} & {\color{refcol}\num {0.770 +- 0.045}} & {\color{refcol}\num {0.765 +- 0.019}} & {\color{refcol}\num {0.436 +- 0.129}} \\ \hline\hline
		{\color{cupcol}\cupnet $(h=\num{7})$} & {\color{cupcol}\num {0.053 +- 0.161}} & {\color{cupcol}\num {0.714 +- 0.191}} & {\color{cupcol}\cnum{0.843 +- 0.049}} & {\color{cupcol}\cnum{0.857 +- 0.033}} & {\color{cupcol}\cnum{0.825 +- 0.051}} & {\color{cupcol}\cnum{0.781 +- 0.049}} \\ \hline       
		{\color{refcol}\regnet $(h=\num{7})$} & {\color{refcol}\rnum{0.772 +- 0.034}} & {\color{refcol}\rnum{0.761 +- 0.058}} & {\color{refcol}\num {0.755 +- 0.040}} & {\color{refcol}\num {0.755 +- 0.031}} & {\color{refcol}\num {0.592 +- 0.243}} & {\color{refcol}\num {0.377 +- 0.203}} \\ \hline
	\end{tabular}
	}%
\end{table}


We find that in most cases, \cupnet is superior to \regnet. It performs worse only for low $\alpha$ and large $h$ (\ie, strong pruning and deep nets). In these cases, the training process appears to not converge to a sufficiently good state. However, the mean scores of the two architectures are mostly within one standard deviation of each other. The best mean \cupnet score of \num{0.862} is achieved for $h=\num{2}$ and $\alpha=\num{5}$, which corresponds to a network with $n_{\mathrm{cup}}=\num{12277950}$ trainable parameters. On the other hand, the best mean \regnet score of \num{0.827} is achieved for $h=\num{3}$ and $\alpha=\num{50}$, which corresponds to a much larger network with $n_{\mathrm{ref}}=\num{34898737}$ trainable parameters. Thus, according to the respective optimal values for $h$ and $\alpha$, our pruning approach leads to smaller networks with a better expected score.\par
As an alternative approach we also test the performance of a Random Forest regressor (with 250 trees) on PCA-transformed features (with 100 components) using the implementation from Ref.~\cite{scikit-learn}. This leads to a $R^2$ score of \num{0.736 +- 0.029} over 10 benchmark runs, which is worse than most results from \cref{tab:results}.

\section{Conclusion}
Summarized, our approach of pruning the network connections according to the spatial correlation of mesh points leads to a better expected performance in comparison with a reference network with similar structure and complexity for our specific use case. A more detailed analysis of this observation is beyond the scope of this paper. However, testing whether our approach can also be applied to other geometrical data could serve as a possible starting point for further research.

\section{Acknowledgements}
We would like to thank Alexander Butz for helpful discussions, Maria Baiker and Jan Pagenkopf for providing the simulation of the cup drawing process, and Boris Giba for data processing. This work was developed in the Fraunhofer Cluster of Excellence \quoting{Cognitive Internet Technologies} as well as the Fraunhofer lighthouse project \quoting{Machine Learning for Production}.

\printbibliography[heading=bibintoc]

\end{document}